\pdfoutput=1

\documentclass[11pt]{article}

\usepackage[]{emnlp2021}

\usepackage{times}
\usepackage{latexsym}

\usepackage[T1]{fontenc}

\usepackage[utf8]{inputenc}

\usepackage{microtype}

\usepackage{amsmath}
\usepackage{booktabs}
\usepackage{color}
\usepackage{graphicx}

\usepackage{array}
\usepackage{makecell} %
\usepackage{amssymb} %
\usepackage{pifont}
\usepackage{multirow}
\usepackage{xurl} %

\usepackage{graphicx} %
\usepackage{subfig} %
\usepackage{booktabs} %
\usepackage{multirow} %
\usepackage{comment} %
\usepackage{amssymb} %
\usepackage{multicol}
\usepackage{dsfont} %

\newcommand{\xmark}{\text{\ding{55}}}

\title{Back to Square One:\\
Artifact Detection, Training and Commonsense Disentanglement\\ in the Winograd Schema}

 \author{Yanai Elazar\textsuperscript{1,2} \,
 Hongming Zhang\textsuperscript{3,4} \,
 Yoav Goldberg\textsuperscript{1,2}\,
 Dan Roth\textsuperscript{4} \\
\textsuperscript{1}Bar Ilan University, \textsuperscript{2}AI2, \textsuperscript{3}HKUST, \textsuperscript{4}UPenn \\
  {\tt  \{yanaiela,yoav.goldberg\}@gmail.com}\\
  {\tt hzhangal@cse.ust.hk, danroth@seas.upenn.edu}\\
  }

\begin{document}
\maketitle

\begin{abstract}

The Winograd Schema (WS) has been proposed as a test for measuring commonsense capabilities of models.
Recently, pre-trained language model-based approaches have boosted performance on some WS benchmarks but the source of improvement is still not clear.  This paper suggests that the apparent progress on WS may not necessarily reflect progress in commonsense reasoning.
To support this claim, we first show that the current evaluation method of WS is sub-optimal and propose a modification that uses twin sentences for evaluation. We also propose two new baselines that indicate the existence of artifacts in WS benchmarks.
We then develop a method for evaluating WS-like sentences in a zero-shot setting to account for the commonsense reasoning abilities acquired during the pretraining and observe that popular language models perform randomly in this setting when using our more strict evaluation. We conclude that the observed progress is mostly due to the use of supervision in training WS models, which is not likely to successfully support all the required commonsense reasoning skills and knowledge.\footnote{The code and evaluation are available at: \url{https://github.com/yanaiela/winograd_square_one}}

\end{abstract}

\section{Introduction}
\label{sec:intro}

The Winograd Schema (WS) \cite{wsc} was proposed as an alternative to the Turing test, by virtue of evaluating progress on commonsense reasoning.
The task is a multi-choice question akin to coreference resolution. Given a text snippet with two entities and a pronoun that refers to one of the entities, select the entity referred to by the pronoun.\footnote{It can also be a possessive adjective, but for simplicity, we refer these as pronouns.}
Consider the following example:

\begin{enumerate}
    \item \label{ex1} \textit{The trophy} doesn't fit into \textit{the brown suitcase} because \textbf{it} is too \underline{large}.
\end{enumerate}
The entities are marked in italics, the pronoun in bold, and the special word\footnote{Words that change the answer. A detailed explanation is provided later.} is underlined. In this case, \textbf{it} refers to \textit{The trophy}, since smaller objects typically fit into larger objects.\footnote{There has been some theoretical work that analyzed WS sentences and proposed a framework, the ``correlation calculus,''  arguing that resolving these problems involves a discourse coherence \cite{bailey2015winograd,michael2015theory}.}

\begin{figure}[t]
    \centering
    \resizebox{\columnwidth}{!}{%

    \begin{tabular}{llr}
    \toprule
       Setup & Example  & Answer \\
         \midrule
        \underline{Original} & & \\
        twin-1 & \makecell[l]{\textit{The trophy} doesn't fit into \textit{the} \\ \textit{brown suitcase} because \textbf{it} is too \underline{large}.} & \raisebox{-3pt}{\includegraphics[width=0.2in]{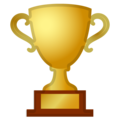}} \enspace trophy \\
        
        twin-2 & \makecell[l]{\textit{The trophy} doesn't fit into \textit{the} \\ \textit{brown suitcase} because \textbf{it} is too \underline{small}.} & \raisebox{-3pt}{\includegraphics[width=0.2in]{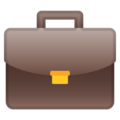}} suitcase \\ \midrule
        
        \underline{Baselines} & & \\
        \textit{no-cands} & \makecell[l]{doesn't fit into because \textbf{it} is too \underline{large}.} & ? \\
        \textit{part-sent} & \makecell[l]{because \textbf{it} is too \underline{large}.} & ?  \\ \midrule
        
        \underline{Zero-shot} & & \\
        twin-1 & \makecell[l]{\textit{The trophy} doesn't fit into \textit{the brown suitcase} \\ because \textbf{the trophy} is too \underline{[MASK]}.} & \raisebox{-6pt}{\includegraphics[width=0.3in]{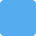}} large  \\
        twin-2 & \makecell[l]{\textit{The trophy} doesn't fit into \textit{the brown suitcase}
        \\ because \textbf{the brown suitcase} is too \underline{[MASK]}.} & \raisebox{-2pt}{\includegraphics[width=0.1in]{figures/blue_square.png}} \enspace small \\

         \bottomrule
    \end{tabular}
    }
    \caption{Examples from the Winograd Schema Challenge (top), our proposed modification to these sentences that we use as novel baselines (middle) and the new formulation of the WS task which allows us to test LMs in a zero-shot setting (bottom).}
    \label{tbl:overview}
\end{figure}

The success of Pretrained Language Models (PLMs) seems to have advanced models' commonsense capabilities by boosting the performance on WS via
simple probability ranking \cite{trinh2018simple,brown2020language,zhou2020evaluating}. 
Another advancement was the curation of a large, crowdsourced dataset for WS, Winogrande \cite{winogrande}.
Models that train on this dataset are close to human performance. But are we any closer to achieving commonsense reasoning?

We provide three explanations for the perceived progress on the WS task: (1) lax evaluation criteria; (2) artifacts in the datasets that remain despite efforts to remove them, and (3) knowledge and reasoning leakage from large training data. 
Combining the effects of these attributes together, we show that all models we consider perform randomly on this task.
Examples for WS, the proposed control baselines, and zero-shot instances can be found in Figure \ref{tbl:overview}.

Our main premise in this work is that, from a commonsense perspective, the generalization capabilities models can get from large training data are limited. Due to the vast number of commonsense facts (e.g. steel is hard, planets are big), it is infeasible to learn them all from a limited-scale training set.
However, this knowledge can still be acquired in different ways, such as self-supervision \cite{NELL-aaai15}, Open IE \cite{tandon2014acquiring}, collecting statistics from large text corpora \cite{elazar-lions}, PLMs \cite{zhou2020evaluating} and more \cite{bagherinezhad2016elephants,forbes2017verb}. Therefore, we claim that the vast majority of commonsense knowledge a model obtains should come from sources external to the supervised dataset. The supervised training set should mainly provide a  means for learning the format of the task but not as a source for commonsense knowledge acquisition.
We thus question the approach, which has recently gained popularity \cite{winogrande,klein-nabi-2020-contrastive}, of using models trained on large datasets for evaluating general commonsense reasoning capabilities, like WS.

\paragraph{Contributions.} (i) We begin by proposing a general evaluation method that makes use of groups that contain similar inputs, e.g. the twin sentences in WS (\S \ref{sec:twin_eval}). That is, instead of measuring accuracy by scoring each sentence separately, we suggest scoring according to the worse score on both inputs: giving a point only if both sentences are predicted correctly.
This evaluation reduces the risk of successful prediction due to artifacts in the data and better reflects the models' commonsense reasoning abilities.
(ii) Next, we extend previous work \cite{trichelair2019reasonable} that manually found in the Winograd Schema Challenge (WSC) associative examples which can be solved using simple statistics. We propose two automatically constructed control baselines that distort the sentences to be nonsensical, on which a score higher than majority suggests the presence of artifacts (\S \ref{sec:biases}).
We find that WSC \cite{wsc} contains a non-trivial amount of artifacts, whereas the newly suggested dataset, WinoGrande \cite{winogrande}, contains much less of these.\footnote{In Appendix \ref{sec:aflite}, we provide details on how \textsc{AfLite}, the algorithm that was used to filter examples from Winogrande operates, and how it is different from our baselines.}

(iii) Finally, to bypass the supervised training step, we propose to directly evaluate PLMs on WS in a zero-shot setup; this allows for assessing how many commonsense reasoning capabilities were acquired in the pretraining step. Specifically, this evaluation disentangles the commonsense capabilities of PLMs from the knowledge they acquire from the training set.
Combining our new evaluation method and taking into account the data artifacts with the zero-shot setting, we show that all models we consider perform randomly.
We then demonstrate using learning curves of models trained on increasing amounts of data, that it takes huge amounts of training instances to make small improvements in the test set, demonstrating the ineffectiveness of large training sets in acquiring commonsense reasoning skills.
We interpret these results as evidence that a lot of the commonsense reasoning capabilities are learned during fine-tuning, as opposed to the pre-training step.

Based on our experiments, we conclude that many of the claims of progress on WS in recent years are unjustified, and stem from sub-optimal evaluation, artifacts, and commonsense knowledge learned from a supervised training set.
Nevertheless, we suggest that the newly proposed Winogrande dataset \cite{winogrande} shouldn't be used for training, but it provides good data for evaluation, and hope that our new evaluation methods will assist faithful tracking of commonsense reasoning progress.

\section{Background}

\subsection{WSC and the Twin Sentences}
\label{sec:background_wsc}

The Winograd Schema Challenge \cite{wsc} was constructed to serve as a benchmark for commonsense reasoning capabilities of models (similarly to the way Textual Entailment was proposed to serve as a benchmark for measuring models' entailment capabilities \cite{dagan2005pascal,DRSZ13}).
WSC contains a small test set of 273 examples, created by experts, and for several years models were struggling to perform well on it.
Each question involves four key features: 1) two entities are mentioned in each sentence, and they can be two males, two females, two inanimate objects, or two groups of people or objects; 2) a pronoun or a possessive adjective is used in the example to refer to one of the entities; 3) the task is to determine which of the two entities is referred to by the pronoun, and 4) each sentence contains a \textit{special word} which, when replaced, the answer changes.
There are no other limitations on the sentences besides these constraints and, consequently, this test is considered to be a general commonsense reasoning test, unlike other benchmarks, which focus on specific commonsense capabilities \cite{rashkin2018event2mind,forbes2019neural,sap2019atomic,sap2019social,bisk2020piqa}.

In order to fulfil the fourth feature, each example was paired with an additional \textit{twin} sentence, which only slightly differs from its twin. (Similar test sets were recently proposed and are referred to as \textit{Counterfactual data} \cite{kaushik2019learning} and \textit{Contrast sets} \cite{contrast-sets}).
For example, the \textit{twin} sentence of Example \ref{ex1} is:

\begin{enumerate}
\setcounter{enumi}{1}
    \item \label{ex2} \textit{The trophy} does not fit into \textit{the brown suitcase} because \textbf{it} is too \underline{small}.
\end{enumerate}
Notice that the special words in these sentences are \textit{large} and \textit{small}, and in this sentence, \textbf{it} refers to \textit{the brown suitcase} (as opposed to \textit{the trophy} in Example \ref{ex1}). The special word is a key part of WS, which makes the task hard to solve. These words were chosen carefully to avoid statistical correlations between the special word and the entities.
In this example, both \textit{trophy} and \textit{suitcase} can be \underline{small}, which makes the task hard to solve by machines; and as \citeauthor{wsc} puts it: ``This helps make the test Google-proof: having access to a large corpus of English text would likely not help much (assuming, that answers to the questions have not yet been posted on the Web, that is)!''

\subsection{Progress on WSC}
\label{sec:backgroun_progress}
Since WSC was proposed as a benchmark for commonsense \cite{wsc}, there were many attempts to improve performance on this benchmark, that involved different approaches including web queries \cite{rahman2012resolving,sharma2015towards,emami2018knowledge}, using external knowledge sources \cite{sharma2019using}, information extraction and reasoning \cite{isaak2016tackling} and more \cite{PengKhRo15,liu2017cause,liu2017combing,fahndrich2018marker,klein-nabi-2019-attention,zhang2019sp,DBLP:conf/www/ZhangLPSL20}.

Newer approaches use LMs to assign a probability to a sentence by replacing the pronoun with an entity, one at a time, and pick the more probable sentence \cite{trinh2018simple,opitz2018addressing,radford2019language,kocijan-etal-2019-surprisingly}. More recently, sequence to sequence models have been employed to directly predict the referred entity in a supervised \cite{raffel2019exploring}, zero-shot or few-shot setting \cite{brown2020language}.
The latest results of GPT-3 \cite{brown2020language} are rather impressive, and agree with the premise of this paper, as the model sees none to a few dozen examples to learn the format. It is worth noting, though, that the training corpus of GPT-3 included some of the WSC questions, and therefore these results should be taken with a grain of salt.
For a comprehensive review of the progress on approaches and related datasets of WS, see \citet{kocijan2020review}.

\citet{zhou2020evaluating} probed multiple LMs for commonsense capabilities in different datasets including WSC, by computing the probability the LM assigns each alternative and choosing the more probable one. The advantage of this method is its unsupervised approach; it does not teach the model any new knowledge.
Notably, their evaluation protocol, which computes the average log probability of each masked word is problematic, since special words that get tokenized into more than one word-piece are still masked independently, thus priming the model towards a certain answer (\S \ref{sec:zs-mlm-eval}). %
In this work, we propose a new evaluation methodology and show that these models' performance is random.
Finally, \citet{winowhy} provided an analysis of different types of commonsense knowledge needed to solve the different WSC questions, including
properties, eventualities, and quantities. 
They also created a new dataset, WinoWhy, which requires models to distinguish between plausible and erroneous reasons for the correct answer.

\section{A Robust Group Score Evaluation %
}
\label{sec:twin_eval}

Many works in recent years have shown that large neural networks can achieve high performance on different benchmarks while ``being right for the wrong reasons'' \cite{mccoy2019right}. These successes arise from a variety of reasons such as artifacts in datasets \cite{poliak2018hypothesis,tsuchiya2018performance,gururangan2018annotation,kaushik2018much}, annotators biases \cite{geva2019annotator}, etc.
\citet{wsc} proposed to alleviate some of these issues by using the twin sentences along with the special word. However, the proposed evaluation of WSC scores each twin separately. As \citet{trichelair2019reasonable} showed that some WSC instances can be solved using simple correlations, we argue that the independent scoring may result in unjustifiably inflated scores. Here, we inspect a new evaluation that accounts for some of these artifacts and provide a more robust evaluation for cases where we have grouped instances (e.g. minimal pairs).

\subsection{Group Scoring}

Recent studies proposed to augment test instances with minimal pairs, that either change the original answer \cite{kaushik2019learning,contrast-sets}, or keep it intact by using paraphrasing, synonyms, etc. \cite{glockner-etal-2018-breaking,shah2019cycle}. Typically, these works report the results separately on the new test set, with no reference to the original test set.

We extend over previous work that proposes to evaluate pairs \cite{abdou-etal-2020-sensitivity} or groups \cite{elazar-consistency} of related instances and assign a point only if they are all correctly predicted by a model.
Our evaluation framework exploits groups of minimal-distance instances and results in a more robust evaluation.
Specifically, for an arbitrary scoring function $f$, and a group of minimal-distance instances $x_i$, score each of the examples $x_{i_j}$ in the group and assign the group its worse-performing score:\footnote{The minimum in cases where higher scores indicate better performance, and maximum otherwise.}
\[
groupScore(x_i) = \min_j f(x_{i_j})
\]

The motivation behind this new evaluation is three-fold: (1) Predicting correctly all examples in a group provides a more robust measurement, and indicates %
a better understanding of the instances; (2) The lowest scored example is the groups' ``Achilles heel'' and thus makes the success on other examples suspicious; (3) It lowers the probability of random predictions (especially in classification tasks), or the use of shallow heuristics to solve examples.
We note that cases where all examples in a group can be solved based on some artifact will still lead to a high score on this group. Therefore this evaluation does not solve the problem of artifacts, but it reduces the chance of scoring them as correct in cases where not all the groups' instances contain artifacts.\footnote{A similar evaluation was used by \citet{ZKNR19}, with the ``Exact Match'' metric for a multi-label classification task.}

In classification tasks, a consequence of this evaluation is the change in random performance. For example, in the case of balanced binary classification, the chance accuracy drops from 50\% to 25\%.

This generic evaluation can be applied not only in classification tasks but also in other tasks that use different evaluation metrics such as BLEU and ROUGE in generation \cite{papineni2002bleu,lin2004rouge}.
For WS, where the task involves a binary classification, we use \textit{group scoring} over the twin sentences, with accuracy as the per-instance scoring function. This yields the paired evaluation that was recently proposed by \citet{abdou-etal-2020-sensitivity} for evaluating WSC. %

\subsection{Other Robust Evaluation Protocols}
\label{sec:optimal-eval}

It is important to note that any WS test set is only an approximation of the commonsense reasoning skills required overall. The twin-sentences allow to test for specific skills (such as the interchange between small and large with `fit' in Examples \ref{ex1}, \ref{ex2}), but other perturbations are possible which allow testing different skills.
For instance, 
\citet{abdou-etal-2020-sensitivity} proposed several perturbations on the original sentences that mostly do not change the answer, such as synonymous entity substitution, tense switch, gender switch, etc.
These perturbations are also reminiscent of the \textit{switched} protocol of \citet{trichelair2019reasonable}, where models are evaluated on examples where the candidates can be switched in the order (which mainly happens with proper names, but also with inanimate objects), expecting a consistent prediction from models since the label does not depend on the entities' order.
Under the \textit{group-scoring} evaluation, we expect a model to succeed on all perturbations from the same group.

\section{Setup}
\label{sec:setup}

\paragraph{Datasets}
We experiment with two English WS datasets: 

Winograd Schema Challenge (WSC) \cite{wsc} contains 273 manually curated examples. We also report results on the \textit{non-associative} examples that were filtered by \citet{trichelair2019reasonable}, named WSC-na.

Winogrande \cite{winogrande} is a recent crowdsourced dataset that contains WS questions. Winogrande contains 40,938, 1,267, 1,767 examples for train, development, and test respectively. Since the test labels were not published, we report our results on the development set.
We provide a more detailed description of these datasets and splits in Appendix \ref{app:setup}.

\paragraph{Modeling}

We follow the modeling of \citet{winogrande}, which finetunes PLMs as a multiple-choice problem on Winogrande's training set. In this modeling, the pronoun is replaced with either one of the entities, and the `[CLS]' token representation is used for prediction. As such, the input format becomes: \texttt{[CLS] context [SEP] entity [SEP]}, which is encoded once which each entity to produce a score.
We also experiment with another loss that was explored in \citet{liu2020precise} where instead of using a different classification head, uses the original MLM head for predictions. We report these results in Appendix \ref{sec:mc-mlm}.

\paragraph{Pre-trained Models}
We experiment with three PLM types: BERT \cite{bert}, RoBERTa \cite{roberta} and ALBERT \cite{albert}.
We provide implementation details in Appendix \ref{app:implementations}.

\section{Artifacts-Detecting Baselines for WS}
\label{sec:biases}

\begin{table}[t]
    \centering
    \begin{tabular}{llcc}
    \toprule
       Dataset & Setup & Single & Group \\
         \midrule
        
        \multirow{3}{6em}{WSC} 
                              & original & 89.71 & 79.41\\
                            
                            & \textit{no-cands} & 60.72 & 40.35\\
                            & \textit{part-sent} & 64.88 & 33.88\\
        
        \midrule
        \multirow{3}{6em}{WSC-na} 
                              & original & 89.45 & 79.09 \\
                            
                            & \textit{no-cands} & 58.06 & 34.41\\
                            & \textit{part-sent} & 59.90 & 25.00\\
        \midrule
        
        \multirow{3}{6em}{Winogrande} 
                            & original & 71.49 & 58.45 \\
                            & \textit{no-cands} & 53.07 & 31.05 \\
                            & \textit{part-sent} & 53.11 & 22.34 \\
        
         \bottomrule
    \end{tabular}
    
    \caption{Results of RoBERTa-large trained on Winogrande, evaluated on the different datasets in the regular condition (original) and the two bias-exposing baselines. Reporting results both on the original accuracy (Single), and the group-scoring (Group). Random performance on the single and group-scoring evaluations are 50\% and 25\% respectively. }
    \label{tab:biases}
    \vspace{-3mm}
\end{table}

WSC was carefully designed by human experts to minimize the presence of artifacts. For instance, Example \ref{ex3} is not considered as a good WS example since a \textit{racecar} is more likely to \textit{go fast} rather than a \textit{school bus}.

\begin{enumerate}
\setcounter{enumi}{2}
    \item \label{ex3} The \textit{racecar} zoomed by the \textit{school bus} because \textbf{it} was going so \underline{fast}.
\end{enumerate}

However, such correlations are often easy to miss. As evidence, \citet{trichelair2019reasonable} found 37 sentences to be \textit{associative}, or \textit{non Google-proof}.\footnote{\textit{Google-proof} is an attribute introduced in \citet{wsc} that refers to the strength of a test, and the inability to solve it by having access to a large text corpora.} These examples were labeled manually using crowdsourcing, therefore these are still bound to what non-experts can catch, and subtler cues may be hard to spot. Other correlations may be harder or impossible to detect by humans since they are the result of spurious correlations \cite{tu2020empirical}. These features, which can be learned during pretraining or fine-tuning, may result in successful predictions that do not reflect commonsense reasoning skills.

To account for these artifacts, we propose two \emph{control baselines}, which are likely to achieve random performance with an artifacts-free model. A score above random indicates the presence of artifacts. %

\paragraph{No-Candidate Baseline}
This baseline (\textit{no-cands}) removes the two candidates (entities) from the text. For instance, Example \ref{ex1} will turn into:
``would not fit into because \textbf{it} is too \underline{large}.''

\begin{table*}[t]
    \centering
    \resizebox{\textwidth}{!}{%
    \begin{tabular}{lllll}
    \toprule
       ID & WSC  & \citeauthor{trichelair2019reasonable} & \textit{no-cands} & \textit{part-sent} \\
         \midrule
        2 & The \textit{trophy} doesn't fit into the brown \textit{suitcase} because \textbf{it} is too \underline{large}. & \xmark & \checkmark & \checkmark \\ \midrule
        8 & The \textit{lawyer} asked the \textit{witness} a question, but \textbf{he} was reluctant to \underline{repeat} it. & \xmark & \xmark & \xmark \\ \midrule
        72 & I couldn't put the \textit{pot} on the \textit{shelf} because \textbf{it} was too \underline{tall}. & \checkmark & \xmark & \xmark \\ \midrule
        185 & \makecell[l]{Sam broke both his \textit{ankles} and he's walking with \textit{crutches}.\\ But a month or so from now \textbf{they} should be \underline{unnecessary}.} & \checkmark & \checkmark & \checkmark \\

         \bottomrule
    \end{tabular}
    }
    \caption{Instances from WSC, along with indication if the manual filtering by \citet{trichelair2019reasonable} marked them as associative,
    and whether our proposed baselines predict them correctly using group scoring.}
    \label{tbl:bias-examples}
\end{table*}

\paragraph{Partial-Sentence Baseline}
In this baseline (\textit{part-sent}) we split the sentence into two parts, based on punctuation and discourse markers\footnote{`so', `but', `and', `because', `although', `though', `due', `since', `.', `,', `;', `?'} 
and take only the part containing the pronoun. For instance, Example \ref{ex1} will be transformed into the following:
``because \textbf{it} is too \underline{large}.''
A similar approach was used by \citet{trichelair2019reasonable}, however, they employed annotators to manually indicate whether the partial sentence containing the pronoun is associative to one of the candidates. Alternatively, we use a trained model and inspect the overall score on a dataset. 

We note that these two baselines create nonsensical sentences. Therefore, we expect humans to not be able to properly solve them. Thus, a model that achieves higher than random performance on these baselines over a large enough dataset is suspected to rely on spurious correlations.

These baselines are reminiscent of previous works that used part of the input (e.g. the hypothesis only baseline in NLI), to reveal artifacts in multiple datasets for NLI \cite{poliak2018hypothesis} and reading comprehension \cite{kaushik2018much}.

\subsection{Results}

We retrain the RoBERTa large model from \citet{winogrande} that was trained on Winogrande and report the results using the original and the new group-based evaluations in Table \ref{tab:biases}.
On WSC this model achieves 89.71\% and 79.41\% accuracy, on WSC-na it achieves 89.45\% and 79.09\%, and on the dev set of Winogrande, it achieves 71.49\% and 58.45\% accuracy, respectively.
To make these evaluations comparable, we filter sentences with no twin sentences from Winogrande and the single triplet sentence from WSC, remaining with 568 and 272 instances, respectively (or, 284 and 136 pairs).
The resulting performance on the original Winogrande development set is 78.3\%.\footnote{Compared to 79.3\%, reported by \citet{winogrande}.} The single accuracy score on sentences that have pairs is lower by almost 7 points than the original set, which suggests that the sentences with no pair are easier, and may contain some artifacts.
Next, we highlight the performance difference between the original evaluation and the paired, which dropped by 10.30, 10.36, and 13.04 points for WSC, WSC-na, and Winogrande, respectively.
Finally, the results on our proposed baselines achieve higher performance than the random baseline for WSC, and the \textit{no-cands} baseline on Winogrande. The \textit{no-cands} baseline achieves 40.35\%, 34.41\%, and 31.05\% on WSC, WSC-na, and Winogrande respectively, whereas the \textit{part-sent} baseline achieves 33.88\%, 25.00\%, and 22.34\% accuracy. These results indicate that WSC contains many artifacts (over 15 points above random performance), and even after the manual filtering of \citet{trichelair2019reasonable} some statistical correlations remain. On Winogrande, the no-cands baseline achieves more than 6 points above random, indicating that it contains fewer artifacts than WSC and WSC-na, presumably due to the \textsc{AfLite} algorithm.

\subsection{Qualitative Analysis}
\label{sec:bias_analysis}

In Table \ref{tbl:bias-examples} we inspect some instances from WSC and indicate if the manual filtering by \citet{trichelair2019reasonable} found them to be associative, and whether our proposed baselines predicted them correctly using group scoring.
Although successful predictions may result from chance (though the probability that both baselines correctly predicted both pairs is relatively low - 6.25\%), we highlight some cases we find interesting.

The first example from the table (ID 2) was predicted correctly by both our baselines, but not by \citet{trichelair2019reasonable}. This may be a case of memorization of this very popular example, by the pretrained RoBERTa model which was trained on many web pages \cite{emami2020analysis}.
We provide some evidence for this example's memorization in Appendix \ref{sec:elaborate_analysis}.
Examples ID 8 and 72 were both predicted incorrectly by our baselines. While the latter was marked as associative by \citet{trichelair2019reasonable}, our baselines did not predict it correctly, perhaps for a good reason; since both a \textit{pot} and a \textit{shelf} can be tall, there's no clear association in this example.
Example ID 185 was predicted correctly by our baselines, as well as by \citet{trichelair2019reasonable} since this example is associative: the word `unnecessary' is more likely to be correlated with \textit{crutches}, rather than \textit{ankles}.

\section{Disentanglement of Commonsense Reasoning and Learned Commonsense}
\label{sec:zero-shot}

\begin{table*}[t]
    \centering
    \resizebox{0.65\textwidth}{!}{%
    \begin{tabular}{l|rr|rr|rr}
    \toprule
    \multirow{2}{*}{Model} & \multicolumn{2}{c|}{WSC} & \multicolumn{2}{c|}{WSC-na} & \multicolumn{2}{c}{WinoGrande} \\
        & Single & Group & Single & Group & Single & Group \\
         \midrule
         random  & 50.00 & 25.00 & 50.00 & 25.00 & 50.00 & 25.00 \\
         \midrule
            BERT-base & 56.52 & 15.22 & 54.79 & 12.33 & 53.12 & 11.11 \\
            BERT-large & 61.41 & 23.91 & 60.27 & 21.92 & 55.56 & 12.50 \\
            RoBERTa-base & 63.04 & 27.17 & 60.27 & 21.92 & 56.25 & 14.58 \\
            RoBERTa-large & 73.91 & 47.83 & 71.23 & 42.47 & 54.86 & 12.50 \\
            ALBERT-base & 55.43 & 13.04 & 55.48 & 12.33 & 52.78 & 7.64 \\
            ALBERT-xxlarge & 78.80 & 57.61 & 77.40 & 54.79 & 58.68 & 20.83 \\
        
         \bottomrule
    \end{tabular}
    }
    \caption{Performance of different PLMs evaluated in the zero-shot setup of WS. Single refers to the standard accuracy over the entire test set, Group refers to group-scoring.}
    \label{tab:zero-shot}
\end{table*}

In this section, we wish to disentangle the commonsense reasoning skills acquired by PLMs during pretraining, and what they learn during fine-tuning on a WS dataset.
We propose a method that allows evaluating pretrained Masked Language Models (MLM) in a zero-shot setting on WS-like questions.

\subsection{Zero Shot MLM Evaluation}
\label{sec:zs-mlm-eval}
Previous work proposed to evaluate MLMs in a zero-shot setting by replacing the \textit{pronoun} with masked tokens, corresponding to the number of tokens the entities are tokenized into. Then, by inspecting each entity's probability the more probable entity is selected \cite{kocijan-etal-2019-surprisingly,abdou-etal-2020-sensitivity}. However, this approach is problematic when the entities are of different token lengths or consist of more than a single token since the model may be primed towards a certain answer. 
For instance, consider Example \ref{ex1}'s entities, trophy and suitcase, in the case they are tokenized into \textit{trophy} and \textit{suit}, \textit{case}. In this scenario, the MLM will see a single mask in one case (and estimate the probability of \textit{trophy}), but in the other case, it will see two masks (assigning the \textit{suit} and \textit{case} probabilities). Since the model has access to the number of tokens it has to complete, the comparison between these two options is flawed.
Another approach, used by \citet{zhou2020evaluating} is to calculate the probability of the entire sentence, by masking a single token at a time. However, this method is also problematic when the entities are tokenized into more than a single token since unmasked tokens are affecting the prediction of the masked tokens.
For instance, following the same example as before, where \textit{suitcase} is tokenized into \textit{suit} and \textit{case}, a model that sees \textit{suit} is more likely to assign a high probability to \textit{case}, therefore staining the probability distribution, and causing a wrong comparison.

Since properly evaluating MLM on WS sentences with more than a single word that differs between the sentences is challenging, we filter these examples.
Then, we mask this word, and compare the probabilities of the two candidates, as was done in previous work \cite{goldberg2019assessing,talmor2019olmpics,ettinger2020bert}.
The issue with this approach is that typically, the candidates are tokenized into multiple word-pieces, which will result in filtering a great portion of the data.
Instead, we propose to make use of the \textit{special} word (the word that is different between the twin sentences), mask it, and replace the pronoun with the correct answer. Then, the model has to decide which of the special words refers to each entity.
Occasionally, there is more than one special word, or it gets tokenized into multiple tokens, therefore we discard these sentences.
An example of this transformation process on Example \ref{ex1} is the following:
\begin{enumerate}
\setcounter{enumi}{3}
    \item \label{ex1-transformed} \textit{The trophy} would not fit into \textit{the brown suitcase} because \underline{the trophy} is too \textbf{[MASK]}.
\end{enumerate}
where `\textbf{[MASK]}' is the token that has to be predicted between the two original special words: `large' or `small'. The twin sentence of this example would accordingly be the same but with the entity `the trophy' replaced with `the brown suitcase', and the correct answer would change from `large' to `small'.

One potential pitfall of this formulation is that it is not faithful to the original WS, and tests a different mechanism.
To test the difference between these formulations, we train the RoBERTa large model on Winogrande on our transformed Winogrande data, and compare it to the results of the same model, trained on the original setup. We make sure to only use sentences that can be transformed, assuring to train both models on the same subset.
The model's performance on the original setup achieves 66.10\% and 55.93\% on the original and paired evaluation development set, whereas the model trained on the transformed setup achieves 70.06\% and 64.97\%. The latter achieves higher performance, suggesting that our transformation may be preferable in modeling, or easier than the original setup. Since this modeling is easier for the model, the results provide a higher bound of the original results, making the following results even more alarming.

We transform WSC, WSC-na, and the Winogrande dev set with the proposed method and remain with 226, 180, and 354 examples, respectively.
We then evaluate the pre-trained LMs described in Section \ref{sec:setup},
and report the results in Table \ref{tab:zero-shot}.
We note that the overall performance is much lower compared to the finetuned model, as expected. Next, the performance on the group-scoring on WSC-na is relatively low, except for RoBERTa-large and ALBERT-xxlarge, which achieve 42.47 and 54.79, high above random performance.
On the other hand, the performance on Winogrande, across all models is below random performance (best result by ALBERT-xxlarge, of 20.83\%), indicating poor commonsense capabilities of these models. Since we found in the previous Section (\S \ref{sec:biases}) that WSC and WSC-na have many artifacts, we take the results on Winogrande to better reflect commonsense reasoning skills.
Recall that the comparison between the two formulations suggested that our new formulation should perform better, a fact that makes the random predictions in the zero-shot setup even more remarkable.

\section{Progress in Commonsense Reasoning?}
The large performance gap may not seem surprising. In most tasks in NLP, we do not expect a PLM to do well on new tasks out of the box and expect a supervised dataset to provide the required skills.
However, we claim that for commonsense tasks, this argument does not hold.
Since commonsense reasoning skills and knowledge are huge, it is not likely to acquire all that information through supervision.
Consider the following WS instances:

\begin{enumerate}
\setcounter{enumi}{4}
    \item \label{ex6} \textit{The large ball} crashed right through the \textit{table} because \textbf{it} was made of \underline{steel}.
    
    \item \label{ex7} I bought a \textit{steel} property at the same time  as my \textit{wooden} property. The \textbf{\_} property was \underline{harder}.
\end{enumerate}
Examples \ref{ex6} and \ref{ex7}%
\footnote{Winogrande was collected with `\_' instead of pronouns.}
come from WSC and Winogrande training set, respectively.
The fact that steel is a strong material is part of the knowledge needed to solve Example \ref{ex6}. However, a model that is trained on Example \ref{ex7} may pick up this fact. Will this training instance also teach the model facts about other materials, such as \textit{styrofoam}?

\begin{figure}[t!]
\centering

\includegraphics[width=1.\columnwidth]{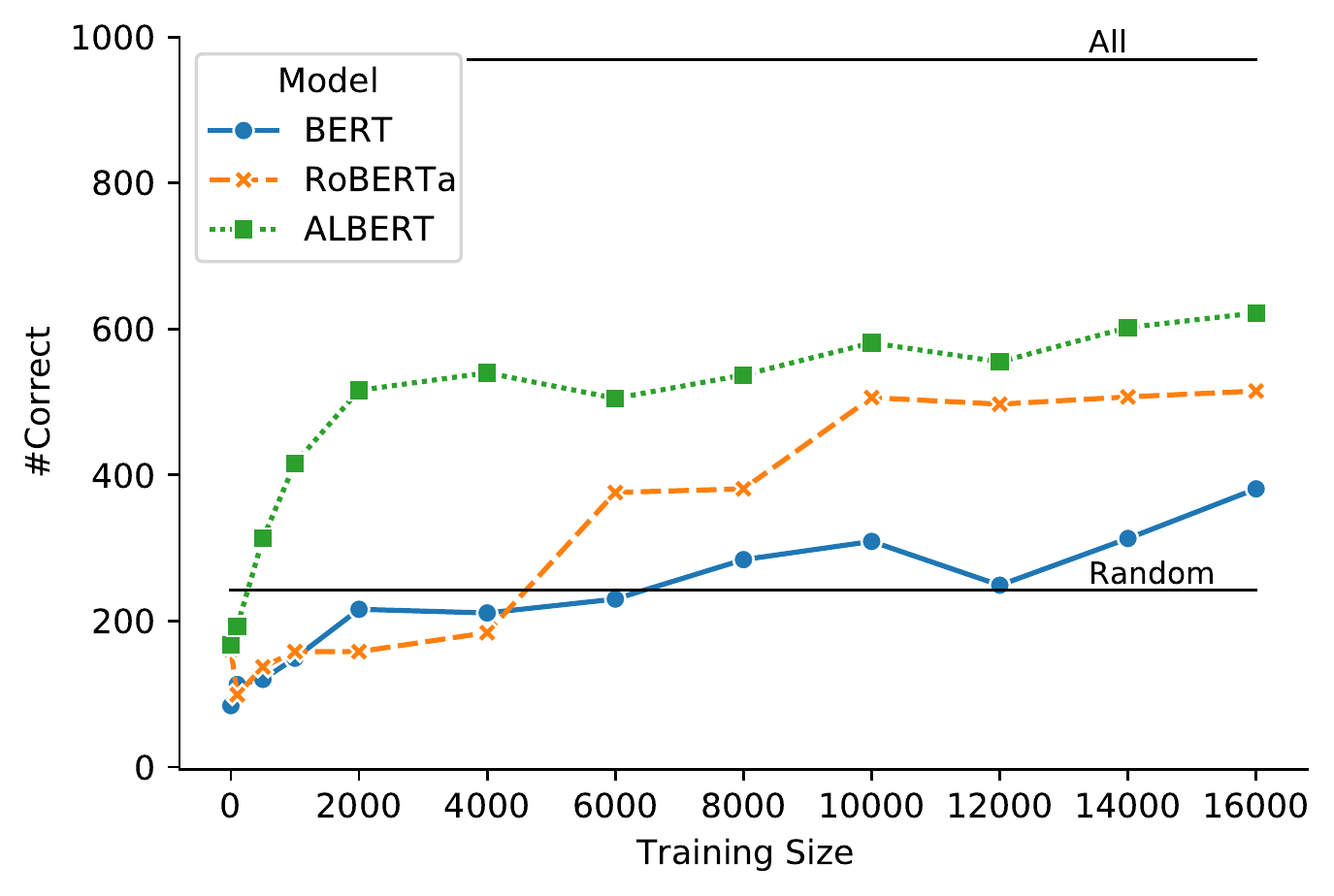}

\caption{Learning curves for the large versions of BERT, RoBERTa, and ALBERT models, trained on increasing amounts of data. 
This figure differs from the Winogrande leaderboard. We explain the source of these differences in Appendix \ref{sec:leaderboard}.
}
\label{fig:learning-curve}
\end{figure}

To quantify the effect of training on the success in solving WS questions, we re-split Winogrande training set to leave enough data for testing and use the rest for training. From the remaining training set, we create multiple training splits, increasing in size, to study the effect of increasing amounts of data on the overall performance. We use the original development set to pick the best models.
We report learning curves with the different models, where each point is the average score of three runs, in Figure \ref{fig:learning-curve}.\footnote{Full numeric results, along with standard deviations are reported in Appendix \ref{sec:app-learning-curves}.} We report the number of correct pairs predicted correctly on the y-axis as a function of the training size.
These curves indicate that the inspected models obtained no commonsense reasoning capabilities in the pretraining step, and are slowly improving their performance the more data they are trained on.
However, except for a sudden improvement with 500 examples for ALBERT, the slope increases incredibly slowly and requires a significant amount of additional training instances for small improvements (BERT and RoBERTa's slopes are more moderate).
We conclude that training data is mostly non-beneficial for generalize commonsense reasoning, and models should acquire it using other methods.

We note that the initial fast increase in ALBERT's performance is interesting, and may be due to another explanation; that is commonsense reasoning is composed of commonsense knowledge (e.g. steel is hard), and reasoning (comparing between objects sizes). Some of the knowledge may be encoded in these models, and reasoning can be taught. However, if that's the case, datasets should account for that, with careful splits. We leave the answer to this question to future work.
Overall, this increase is nevertheless rather moderate, and once a model passes this point (about 2000 examples), the performance increases slowly, which goes in line with our claims.

A potential explanation for the sudden performance improvement with finetuning, and the lower baselines scores on Winogrande, may arise from the unnaturalness aspect of this dataset. For instance, we find Example \ref{ex6} from WSC a more natural sentence than Example \ref{ex7} (from Winogrande). Thus, in the case of several less-natural occurring sentences in Winogrande, the random results of our baselines may be explained due to this fact, and the finetuning procedure may contribute to the model's adaptation of that language. We leave the assessment of this hypothesis to future work.

\section{Conclusions}
\label{sec:discussion}

In this work, we begin by discussing the current evaluation of WS and propose an additional evaluation metric, \textit{group-scoring}, that credits a model with the worse performing instance of a group.
While we focus here on WS, we propose to use this evaluation in other tasks, where minimal pairs are available \cite{kaushik2019learning,contrast-sets,warstadt2020blimp}, as a more reliable evaluation metric.
We then propose two new control baselines that account for artifacts in WS data and show that WSC contains many artifacts, while Winogrande consists much less of them.

Finally, we propose a method to evaluate MLMs on WS sentences in a zero-shot setting. We show that the performance of popular MLMs is random and that models improve gradually the more training data they see.
We conclude that the use of large training sets is not always desirable, especially in commonsense reasoning settings, and call future work to find other methods to improve our models' commonsense abilities.

\section*{Acknowledgements}
We would like to thank Vered Shwartz, Keisuke Sakaguchi, Rotem Dror, Niket Tandon, Vid Kocijan and Ernest Davis for helpful discussions and comments on early versions of this paper.
We also thank the anonymous reviewers for their valuable suggestions.

Yanai Elazar is grateful to be supported by the PBC fellowship for outstanding PhD candidates in Data Science and the Google PhD fellowship.
This project has received funding from the Europoean Research Council (ERC) under the Europoean Union's Horizon 2020 research and innovation programme, grant agreement No. 802774 (iEXTRACT) and from contract FA8750-19-2-1004 with the US Defense Advanced Research Projects Agency (DARPA).

\bibliography{custom}
\bibliographystyle{acl_natbib}

\newpage
\clearpage

\appendix

\section{Detailed Setup}
\label{app:setup}
\paragraph{Datasets}
We report our results on two datasets: 

Winograd Schema Challenge (WSC) \cite{wsc} contains 273 manually curated examples.
Each example is paired with a \textit{twin-sentence}, meaning that there's a special word that is changed between the two sentences, that changes the coreferring entity.
\citet{trichelair2019reasonable} have labeled the original WSC examples, and found 37 examples to be \textit{associative} \citet{trichelair2019reasonable}. We thus also use the \textit{non-associative} subset which excludes the associative examples. We refer to this subset as WSC-na

Winogrande \cite{winogrande} is a recent crowdsourced dataset that contains WS questions. Winogrande is much larger than WSC and contains 9,248, 1,267, 1,767 examples for train, development, and test respectively.
Winogrande was filtered from `biases' (or artifacts) using their proposed \textsc{AfLite} algorithm, which produced the mentioned challenging dataset. However, the authors also release and use the `biased' instances for training, making a total of 40,938 training instances.

\paragraph{Pre-trained Models}
We experiments with multiple pre-trained models: BERT \cite{bert}, RoBERTa \cite{roberta} and ALBERT \cite{albert}.
These models are large Transformer-based architectures \cite{vaswani2017attention}, that are trained on the Masked Language Modeling task, which is predicting the masked word in a given context.
These models are pretrained on huge amounts of text such as Wikipedia, the book corpus \cite{book-corpus}, parts of CommonCrawl, and more.
Specifically, we conduct our experiments with BERT-large-cased, RoBERT-Large, and ALBERT-XXLarge-V2, which have 335M, 335M, and 223M parameters, respectively.

\section{Implementation Details}
\label{app:implementations}
We implemented the experiments with the huggingface package~\cite{wolf-etal-2020-transformers}.
Following the previous work~\cite{winogrande}, on all our experiments, we set the learning rate to be 1e-5, batch size to be 8, and trained the models for 8 epochs.
Adam~\cite{DBLP:journals/corr/KingmaB14} is used as the optimizer.
We optimize all models with the cross-entropy loss function.
We trained our model with RTX 2080, and the training time is 13, 14, and 62 minutes per epoch on the largest training set Winogrande (10) for BERT-large, RoBERTa-large, and Albert-XXL-v2, respectively. 
As the evaluation is conducted on the dev set, we do not use it to select the best model. 
Instead, we report the performance with the final model, which is converged based on our observation.

\section{Full Learning Curves Results}
\label{sec:app-learning-curves}

The full results from Figure \ref{fig:learning-curve}, along with the standard deviations, are reported in Table \ref{tab:learning-curves-all}.

\begin{table*}[t]
    \centering
    \begin{tabular}{r|rr|rr|rr}
    \toprule
    
    \multirow{2}{*}{\# Training} & \multicolumn{2}{c|}{BERT} & \multicolumn{2}{c|}{RoBERTa} & \multicolumn{2}{c}{ALBERT} \\
        & Single & Group & Single & Group & Single & Group \\
         \midrule
        0   & 52.99 (0.00) & 8.67 (0.00) & 56.39 (0.00) & 16.61 (0.00) & 55.55 (0.00) & 17.23 (0.00) \\
        100 & 53.47 (0.75) & 11.71 (0.75) & 52.78 (0.75) & 10.22 (4.48) & 58.24 (1.49) & 19.89 (2.74) \\
        500 & 49.31 (1.87) & 12.42 (1.74) & 49.65 (0.50) & 14.17 (1.99)  & 60.07 (0.50) & 32.35 (1.27) \\
        1,000 & 51.39 (0.99) & 15.33 (0.75) & 50.35 (0.37) & 16.33 (0.50)  & 62.50 (0.62) & 42.89 (0.25)\\
        2,000 & 51.39 (0.87) & 22.32 (3.49) & 49.65 (0.25) & 16.35 (1.49)& 62.85 (2.36) & 53.27 (3.24) \\
        4,000 & 48.96 (2.61) & 21.73 (2.49) & 49.65 (0.50) & 18.94 (1.24) & 67.36 (1.12) & 55.72 (2.49)\\
        6,000 & 50.35 (0.50) & 23.73 (0.50) & 59.72 (1.86) & 38.85 (2.99) & 67.71 (2.86) & 52.16 (3.73) \\
        8,000 & 48.26 (1.24) & 29.27 (0.75) & 50.35 (1.37) & 39.32 (1.99) & 67.36 (0.50) & 55.43 (0.25) \\
        10,000 & 51.39 (1.12) & 31.85 (1.99)& 62.85 (0.12) &  52.27 (0.99) & 73.76 (1.76) & 59.98 (1.94) \\
        12,000 & 50.00 (1.62) & 25.68 (1.49) & 62.85 (0.50) & 51.24 (0.50) & 72.22 (1.33) & 57.28 (0.54) \\
        14,000 & 52.08 (0.50) & 32.31 (3.24) & 62.15 (1.49) & 52.31 (0.75)  & 75.61 (0.63) & 62.15 (2.24) \\
        16,000 & 54.86 (0.75) & 39.31 (1.99) & 60.42 (2.11) &  53.14 (3.24) & 76.82 (1.15) & 64.21 (1.42) \\ 
         \bottomrule
    \end{tabular}
    \caption{Effect of the training data size on different models performance. We report results on BERT, RoBERTa and ALBERT, all with their largest variants.}
    \label{tab:learning-curves-all}
\end{table*}

\section{\textsc{AfLite} Details}
\label{sec:aflite}

\textsc{AfLite} \cite{winogrande}, an algorithm proposed for reducing datasets' artifacts was used to create Winogrande \cite{winogrande}. It works as follows: a RoBERTa model \cite{roberta} is finetuned on a random subset of the data to train a `weak' model of the task. Then, the rest of the instances are encoded using the model's encoder. Then, for multiple iterations, a set of weak classifiers (linear) are trained on a subset of the encoded data and predict the rest. If more than k classifier predicted correctly an instance’s label, it is discarded from the final dataset. This process is repeated multiple times until reaching a satisfying dataset size (which is controlled by predefined hyperparameters).

Although this algorithm filter examples that are `easy', as a set of linear models that were trained on a medium quality representation managed to predict the correct answer, it is unclear how artifact-free the dataset is.
In contrast, our proposed baseline methods directly detect artifacts the classification model may rely on, by presenting challenging perturbations on which a model is not likely to succeed above random.
Thus, our procedure is inherently different than the general-purpose \textsc{AfLite} filtering algorithm.

\section{Comparison to Winogrande Leaderboard}
\label{sec:leaderboard}
We note that Figure \ref{fig:learning-curve} differs from the \href{https://leaderboard.allenai.org/winogrande/submissions/public}{Winogrande leaderboard} in multiple ways: first, we compare different models than the ones that appear on the leaderboard. Specifically, the to-date leading submission (accurate as of March 21st, 2021), UNICORN, does not provide details about the model, except it is a T5 based model, trained on a collection of datasets. Since the content of these datasets is not publicly available, it is impossible to assess the quality of this submission. For instance, if one of these datasets contains other commonsense reasoning datasets, the model may have picked up on commonsense reasoning skills which are also tested for in Winogrande.
Second, the leaderboard uses the original evaluation, based on the accuracy of single instances. As we claim in Section \ref{sec:twin_eval}, this evaluation is sub-optimal and causes an over-estimation of the actual performance of models.
Moreover, our analyses were done on the development set, as opposed to the reported test set performance, since the test set is not publicly available.
Finally, the leaderboard presents a learning curve of 5 training sizes, as we report the results over 12 different training sizes.

\section{Elaborate Analysis}
\label{sec:elaborate_analysis}
In Section \ref{sec:bias_analysis} we showcase some examples from WSC and provide possible explanations for which our baselines (\S \ref{sec:biases}) are able to solve them.
Here, we provide additional evidence that supports our claim. We do so for the example where both baselines predict the correct answer, but the manual inspection from \citet{trichelair2019reasonable} does not consider it to be associative. We emphasize that this example is not associative per se, and thus the annotation from \citet{trichelair2019reasonable} was correct, but the pretrained model, which was trained on the web, may have caught up statistical cues that help it predict these examples correctly, even with partial information.
For completeness, we repeat the example here:

\begin{enumerate}
\setcounter{enumi}{6}
    \item \label{ex8} The \textit{trophy} doesn't fit into the brown \textit{suitcase} because \textbf{it} is too \underline{large}.
    
\end{enumerate}

Example \ref{ex8} is a popular example that is often given when describing the task in the media. As evidence, we search for this sentence in Google and found it in multiple websites:
\begin{itemize}
    \item \url{https://theness.com/neurologicablog/index.php/a-tougher-turing-test/}
    \item \url{https://www.eitdigital.eu/newsroom/blog/article/whats-too-big-the-trophy-or-the-suitcase/}
    \item \url{https://cmte.ieee.org/futuredirections/2014/08/20/whats-too-big-the-trophy-or-the-suitcase/}
\end{itemize}

Next, we search for these websites in Common Crawl\footnote{\url{https://commoncrawl.org/}}, the February 2019 version that was reported to be part of RoBERTa's training data \cite{roberta}.
We use an index server\footnote{\url{http://index.commoncrawl.org/CC-MAIN-2019-09/}} that allows querying a specific index and look specific websites. We find that the first two websites are included in this index.
Although we cannot guarantee that these websites were part of RoBERTa's training data since it was not published, the probability that several examples from WSC were part of the large training data of RoBERTa (and later models), with these websites, or other, is high.

\begin{table*}[t]
    \centering
    \begin{tabular}{r|rr|rr|rr}
    \toprule
    
    \multirow{2}{*}{\# Training} & \multicolumn{2}{c|}{BERT} & \multicolumn{2}{c|}{RoBERTa} & \multicolumn{2}{c}{ALBERT} \\
        & Single & Group & Single & Group & Single & Group \\
         \midrule
        0   & 52.99 (0.00) & 8.67 (0.00)  & 56.39 (0.00) & 16.61 (0.00) & 55.55 (0.00) & 17.23 (0.00) \\
        100 & 54.39 (1.59) & 12.28 (2.39)& 55.46 (0.18) & 17.61 (1.46) & 56.14 (1.12) & 17.54 (3.73) \\
        500 & 51.32 (0.37) & 10.53 (2.14) & 55.63  (1.67) & 25.00  (3.27) & 61.97 (1.37) & 34.15 (1.74) \\
        1,000 & 51.75 (0.63) & 12.28 (0.89) & 58.27 (2.03) & 35.56 (3.27) & 62.85 (0.12) & 34.86 (0.25) \\
        2,000 & 54.93 (0.37) & 14.44 (1.33) & 57.92 (1.09) & 35.21 (3.16) & 61.44 (0.75) & 34.51 (0.50) \\
        4,000 & 52.46 (0.71) & 16.55 (2.00) & 61.09 (1.84) & 37.32 (2.67) & 64.08 (2.49)& 40.49 (2.32) \\
        6,000 & 53.87 (1.57) & 20.07 (1.53) & 59.15 (0.62) & 37.32 (1.51) & 68.66 (1.12) & 49.65 (0.75)\\
        8,000 & 53.69 (1.17) & 22.89 (1.08) & 62.15 (0.98) & 39.44 (1.27) & 68.13 (2.74) & 50.70 (3.73)\\
        10,000 & 53.87 (0.51) & 25.00 (1.61) & 63.56 (1.24) & 45.77 (2.21) & 70.42 (1.49) & 53.17 (0.50) \\
        12,000 & 50.17 (1.50) & 23.94 (2.46) & 64.26 (1.07) & 45.42 (3.27) & 69.54 (0.51) & 52.46 (0.50)\\
        14,000 & 52.82 (2.00) & 27.11 (3.86) & 63.38 (1.25) & 44.72 (2.99) & 67.61 (0.97) & 53.17 (1.81)\\
        16,000 & 53.69 (0.67) & 27.11 (0.89) & 61.09 (0.57) & 41.67 (1.77) & 70.77 (0.75) & 55.28 (1.23) \\
         \bottomrule
    \end{tabular}
    \caption{Effect of the training data size on different models performance. We report results on BERT, RoBERTa and ALBERT, all with their largest variants.}
    \label{tab:learning-curves-all-mc-mlm}
\end{table*}

\begin{table}[t]
    \centering
    \begin{tabular}{llcc}
    \toprule
      Dataset & Setup & Single & Group \\
         \midrule
        
        \multirow{3}{6em}{WSC} 
                              & original & 89.71 & 80.88 \\
                            
                            & \textit{no-cands} & 60.96 & 29.82 \\
                            & \textit{part-sent} & 59.09 & 22.31\\
        
        \midrule
        \multirow{3}{6em}{WSC-na} 
                              & original & 90.00 & 81.82 \\
                            
                            & \textit{no-cands} & 59.14 & 25.81\\
                            & \textit{part-sent} & 56.77 & 16.67\\
        \midrule
        
        \multirow{3}{6em}{Winogrande} 
                            & original & 70.95 & 54.23 \\
                            & \textit{no-cands} & 54.87 & 17.69 \\
                            & \textit{part-sent} & 54.43 & 14.18 \\
        
         \bottomrule
    \end{tabular}
    
    \caption{Results of RoBERTa-large trained on Winogrande, evaluated on the different datasets in the regular condition (original) and the two bias-exposing baselines using the MC-MLM loss \cite{liu2020precise}. Reporting results both on the original accuracy (Single), and the group-scoring (Group). Random performance on the single and group-scoring evaluations are 50\% and 25\% respectively.}
    \label{tab:biases-mc-mlm}
    \vspace{-3mm}
\end{table}

\section{MLM results}
\label{sec:mc-mlm}

Here we report the results for the MC-MLM loss that was explored in \citet{liu2020precise}, where instead of training a new head for the classification task, it uses the original MLM head and scores the different candidates instead of the pronoun.
We run all experiments including fine-tuning, and report the results in this section.

The artifacts experiment results are detailed in Table \ref{tab:biases-mc-mlm}. Although the results on the standard setting (\textit{original}) are similar to the ones when using a dedicated head (Table \ref{tab:biases}), this model appears to rely less on artifacts: the \textit{no-cands} baseline still perform better than random on WSC, but the other baseline and the other evaluations perform randomly.

Finally, we repeat the learning curves experiment using the MC-MLM loss, on increasing amounts of data, where for each training size we train 3 models and report the mean and std, and report the results in Table \ref{tab:learning-curves-all-mc-mlm}.
Here, in contrast to the trends shown in \citet{liu2020precise}, we observe generally worse results using the MC-MLM loss. One source of difference is that \citet{liu2020precise} repeated the experiments many more times while performing a grid search over different hyperparameters, while we used the same default hyperparameters for all experiments. Another source of difference is the different training and evaluation splits used in our studies.
We conclude that nevertheless, the trends remain the same, and the slopes of both methods are slow to increase, and thus strengthens our claims about the limited usefulness of training data for WS.

\end{document}